\documentclass{article}

\usepackage[nonatbib, final]{neurips_2019}
\usepackage{style_extra}
 \usepackage[numbers]{natbib}

\usepackage[hidelinks, colorlinks=false]{hyperref}

\title{Increasing Expressivity of a Hyperspherical VAE}

\author{%
  Tim R. Davidson \\
  Aiconic \\
  \texttt{tim.davidson@aiconic.com} \\
  \And
  Jakub M. Tomczak \\
  University of Amsterdam \\
  \texttt{jmk.tomczak@gmail.com} \\
  \And
  Efstratios Gavves \\
   University of Amsterdam \\
  \texttt{egavves@uva.nl} \\
}

\begin{document}

\maketitle

\begin{abstract}
    Learning suitable latent representations for observed, high-dimensional data is an important research topic underlying many recent advances in machine learning.
    While traditionally the Gaussian normal distribution has been the go-to latent parameterization, recently a variety of works have successfully proposed the use of manifold-valued latents. 
    In one such work \citep{s-vae}, the authors empirically show the potential benefits of using a hyperspherical von Mises-Fisher (vMF) distribution in low dimensionality. 
    However, due to the unique distributional form of the vMF, expressivity in higher dimensional space is limited as a result of its scalar concentration parameter leading to a `hyperspherical bottleneck'. In this work we propose to extend the usability of hyperspherical parameterizations to higher dimensions using a product-space instead, showing improved results on a selection of image datasets.
\end{abstract}

\section{Introduction}

Following the manifold hypothesis, in unsupervised generative learning we strive to recover a distribution on a (low-dimensional) latent manifold, capable of explaining observed, high-dimensional data, e.g. images. One of the most popular frameworks to achieve this goal is the Variational Auto-Encoder (VAE) \citep{kingma-vae13, pmlr-v32-rezende14}, a latent variable model which combines variational inference and auto-encoding to directly optimize the parameters of some latent distribution. While originally restricted to `flat' space using the classic Gaussian normal distribution, there has recently been a surge in research extending the VAE to distributions defined on manifolds with non-trivial topologies \citep{s-vae, falorsi2018explorations, mathieu2019hierarchical, hyperbolic-exp-map-reparam19, diffusion-vae19, implicit_reparam_figurnov, relie-aistats19}. This is fruitful, as most data is not best represented by distributions on flat space, which can lead to undesired `manifold mismatch' behavior. 

In \citep{s-vae}, the authors propose a hyperspherical parameterization of the VAE using a von Mises-Fisher distribution, demonstrating the improved results over the especially bad pairing of the `blob-like' Gaussian distribution and hyperspherical data. Surprisingly, they further show that these positive results extend to datasets \emph{without} a clear hyperspherical interpretation, which they mostly attribute to the restricted surface area and the absence of a `mean-biased' prior in the vMF as the Uniform distribution is feasible in the compact, hyperspherical space. However, as dimensionality increases performance begins to decrease. This could possibly be explained by taking a closer look at the vMF's functional form
\begin{align}
    q(\z|\mathbf{\mu}, \kappa) &=  \mathcal{C}_m(\kappa)\exp{(\kappa \mathbf{\mu}^T\z)}, \label{eq2:vmf-def} \\
    \mathcal{C}_m(\kappa) &= \dfrac{\kappa^{m/2 - 1}}{(2\pi)^{m/2}\mathcal{I}_{m/2 - 1}(\kappa)}, \label{eq2:normalizing-constant}
\end{align}
where $||\mathbf{\mu}||^2 = 1$, $\kappa$ a scalar, $\mathcal{C}_m(\kappa)$ is the normalizing constant, and $\mathcal{I}_v$ denotes the modified Bessel function of the first kind at order $v$. Note that the scalar concentration parameter $\kappa$ is fixed in all dimensions, severely limiting the distribution's expressiveness as dimensionality increases.

\section{Method: A Hyperspherical Product-Space}

To improve on the vMF's per-dimension concentration flexibility limitation we propose a simple idea: breaking up the single latent hyperspherical assumption, into a concatenation of multiple independent hyperspherical distributions. Such a compositional construction increases flexibility through the addition of a new concentration parameter for each hypersphere, as well as providing the possibility of sub-structure forming 
. Given a hyperspherical random variable $\z \in \Sm{M}$, we want to choose $\z_0, \z_1, \cdots, \z_k$ in respectively $\Sm{M_0}, \Sm{M_1}, \cdots, \Sm{M_k}$ s.t. $\sum_{i=0}^k M_i = M$, and $\z = \z_0 \frown \z_1 \frown \cdots \frown \z_k$, where $(\frown)$ denotes concatenation. The probabilistic model becomes:
\begin{equation}
    p(\z) = p(\z_0, \z_1, \cdots, \z_k) \stackrel{*}{=} \prod_{i=0}^k p(\z_i), \label{eq:main-prob-dim-permutation}
\end{equation}
which factorizes in (*) if we assume independence between the new sub-structures. Assuming conditional independence of the approximate posterior as well, i.e. $q(\z | \x) = q(\z_0 | \x)q(\z_1 | \x) \cdots q(\z_k | \x)$, it can be shown\footnote{See Appendix \ref{ap:dim-decomp} for derivation.} that the Kullback-Leibler divergence simplifies as
\begin{align}
    KL(q(\z|\x) || p(\z)) &= \int_{\Sm{M}} q(\z|\x) \log \dfrac{q(\z|\x)}{p(\z)} d\z 
    = \sum_i KL(q(\z_i | \x) || p(\z_i)) \label{eq:main-dim-permutation-kl}
\end{align}

\paragraph{Flexibility Trade-Off} \label{par:flexibility-trade-off}
Given a single hypersphere and keeping the ambient space fixed, for each additional `break', a degree of freedom is exchanged for a concentration parameter. In the base case of \Sm{k+1}, we can potentially support $k+1$ `independent' feature dimensions, that must share a single concentration parameter $\kappa$, and hence are globally restricted in their flexibility per dimension. On the other hand, the moment we break \Sm{k+1} up in the Cartesian cross-product of $\Sm{k/2} \times \Sm{k/2}$, we `lose' an independent dimensions (or degree of freedom), but in exchange the two resulting sub-hyperspheres have to share their concentration parameters $\kappa_1, \kappa_2$ over fewer dimensions increasing flexibility\footnote{In the most extreme case, this will lead to a latent space of $[\Sm{1}]_{\times (k+1)/2}$ - which is equal to the $n$-Torus.}.

The reason a vMF is uniquely suited for such a decomposition as opposed to a Gaussian, is that assuming a factorized variance the Gaussian distribution is already equipped with a concentration parameter for each dimension. However, in the case of the vMF, which has only a single concentration parameter $\kappa$ for \emph{all} dimensions, we gain flexibility. This is an important distinction: while all dimensions are implicitly connected through the shared loss objective in both cases, in the case of the vMF this connection is amplified through the \emph{direct connection} of the shared concentration parameter. 

\paragraph{Related Work}
The work closest to our model is that of \citep{paquet2018factorial}, where a Cartesian product of Gaussian Mixture Models (GMMMs) is proposed, with hyperpriors on all separate components to create a fully data-inferred model. They use results from \citep{hoffman2013stochastic,  johnson2016composing} on structured VAEs, and extend the work on VAEs with single GMMs of \citep{nalisnick2016approximate, dilokthanakul2016deep, jiang2017variational}. Partially following similar motivations to our work, the authors hypothesize and empirically show the structured compositionality encourages disentanglement. By working with GMMs instead of single Gaussians, they circumvent the factorized single Gaussian break-up limitation described before. Another recent work proposing to break up a large, single latent representation into a composition of sub-structures in the context of Bayesian optimization is \citep{combo-oh19}.

\section{Experiments and Discussion}

To test the ability of a hyperspherical product-space model to improve performance over its single-shell counterpart, we perform product-space interpolations breaking up a single shell into an increasing amount of independent components. 

\paragraph{Experimental Setup} We conduct experiments on Static MNIST, Omniglot \citep{lake2015human}, and Caltech 101 Silhouettes \citep{marlin2010inductive} mostly following the experimental setup of \citep{s-vae}, using a simple MLP encoder-decoder architecture with \texttt{ReLU()} activations between layers. We train for 300 epochs using early-stopping with a look-ahead of 50 epochs, and a linear \textit{warm-up} scheme of 100 epochs as per \citep{bowman2015, sonderby2016ladder}, during which the KL divergence is annealed from 0 to $\beta$ \citep{higgins2017beta, alemi2018fixing}. Marginal log-likelihood is estimated using importance sampling with 500 sample points per \citep{burda2015importance}, reporting the mean over three random seeds.

Keeping in mind the \emph{flexibility trade-off} consideration, we analyze both the effects of keeping the total degrees of freedom fixed (increasing ambient space dimensionality), as well as the case of keeping the ambient space fixed (decreasing the degrees of freedom). We break up \Sm{40} respectively into 2, 4, 6, 10, and 40 sub-spaces. In each break-up, we try a balanced, leveled, and unbalanced hyperspherical composition.

\paragraph{Results}

\begin{table}[!t]
\centering
\caption{Overview of best results of various \Sm{40} product-space ambient dimensionality interpolations compared to best single \Sm{k}-VAE ($k \le 40$) indicated (*). LL represents the negative log-likelihood, $\mathcal{L}|q|$ the ELBO, $a$ indicates the ambient space dimensionality, $\kappa$ the number of concentration parameters, i.e. breaks, and [\Sm{k}] the product-space composition.
}
\bigskip
  \bgroup
  \setlength\tabcolsep{.5em}
  \begin{tabular}{lcccccccc}
    \toprule
        \multirow[b]{2}{*}{\textbf{$a$}} & & & \multicolumn{5}{c}{Static MNIST} &
        \\
        & $\kappa$ & $[\mathcal{S}_k]$ & LL & $\mathcal{L}|q|$ & & LL* & $\mathcal{L}|q|$* \\
        \midrule
        41 & 4 & $ \Sm{10} \times [\Sm{9}]_{\times 3} $ & -92.65 & -98.23  & & -96.32 & -104.11\\
        41 & 4 & $ \Sm{20 \times 10 \times 6 \times 1} $ & -92.59 & -98.27 & & & \\
        41 & 6 & $ \Sm{15 \times 10 \times 4 \times 3 \times 2 \times 1} $ & -92.25 & -98.10& & & \\
        41 & 6 & $ [\Sm{6}]_{\times 5} \times \Sm{5} $ & -92.71 & -98.46 & & & \\
        \multicolumn{8}{c}{}\\
        & & & \multicolumn{5}{c}{Caltech} &  \\
        \midrule
        41 & 4 & $ \Sm{10} \times [\Sm{9}]_{\times 3} $ & -139.30 & -151.67  & & -143.49 & -152.25 \\
        41 & 4 & $ \Sm{20 \times 10 \times 6 \times 1} $ & -140.64 & -153.05 & & & \\
        \multicolumn{8}{c}{}\\
        & & & \multicolumn{5}{c}{Omniglot} & \\
        \midrule
        41 & 4 & $ \Sm{20 \times 10 \times 6 \times 1} $ & -112.79 & -119.17 & & -113.83 & -120.48 \\
        41 & 6 & $ [\Sm{6}]_{\times 5} \times \Sm{5} $ & -112.58   & -118.49 & & & \\
        41 & 10 & $ \Sm{4} \times [\Sm{3}]_{\times 9} $ & -112.64  & -118.67 & & & \\
    \bottomrule
  \end{tabular}
  \egroup
  \label{tab:mtr-ambient-s40-overview}
\end{table}

A summary of best results for fixed ambient space is shown in Table \ref{tab:mtr-ambient-s40-overview}, with a summary of best results for fixed degrees of freedom and complete interpolations in Appendix \ref{ap:sup-tables-fig}. Initial inspection shows that partially breaking up a single \Sm{40} hypersphere into a hyperspherical product-space indeed allows us to improve performance for all examined datasets. Diving deeper into the results, we do find that both the number of breaks as well as the dimensional composition of these breaks strongly inform performance and learning stability. 

A high number of breaks appears to negatively influence both performance and learning stability. Indeed, for most datasets the `Torus' setting, i.e. full factorization in \Sm{1} components proved too unstable to train to convergence. One explanation for this result could be found in the fact that we omit the REINFORCE part of the vMF reparameterization during training\footnote{See \citep{s-vae}, Appendix D for more details.}. While only of very limited influence on a single hyperspherical distribution, the accumulated bias across many shells might lead to a non-trivial effect. On the other hand, adding as few as four breaks extends the model's expressivity enough to outperform a single shell consistently.

Balance of the subspace composition plays a key role as well. We find that when the subspaces are too unbalanced, e.g. \Sm{37} v. $[\Sm{1}]_{\times 3}$, the network starts to `choose' between subspace channels. Effectively, it will for example start encoding all information in the \Sm{1} shells and \emph{completely ignore} the \Sm{37} shell, leading to an effective latent space of \Sm{3} degrees of freedom\footnote{For a more extended discussion on the interplay between balance and the KL divergence see Appendix \ref{ap:sup-tables-fig}.}, see for example Fig. \ref{fig:s9_1x9_ignored}. On the contrary, better balanced compositions appear capable of cleanly separating semantically meaningful features across shells as displayed in Fig. \ref{fig:s9_1x9_thick_thin}. 

\paragraph{Conclusion and Future Work} In summary, breaking up a single hypersphere into multiple components effectively increases concentration expressiveness leading to more stable training and improved results. In future work we'd like to investigate the possibility of \emph{learning} an optimal break-up as opposed to fixing it a-priori, as well as mixing sub-spaces with different topologies.

\newpage 

\subsubsection*{Acknowledgments}
We would like to thank Luca Falorsi and Nicola De Cao for insightful discussions during the development of this work.


\newpage

\appendix

\section{Dimensionality Decomposition} \label{ap:dim-decomp}
Given a latent variable $\z \in \R^M$, we choose $\z_0, \z_1, \cdots, \z_k$ in respectively $\R^{M_0}, \R^{M_1}, \cdots, \R^{M_k}$ s.t. $\sum_{i=0}^k M_i = M$, and $\z = \z_0 \frown \z_1 \frown \cdots \frown \z_k$, where $(\frown)$ denotes concatenation. The probabilistic model becomes:
\begin{equation}
    p(\z) = p(\z_0, \z_1, \cdots, \z_k) \stackrel{*}{=} \prod_{i=0}^k p(\z_i), \label{eq:prob-dim-permutation}
\end{equation}
which factorizes in (*) if we assume independence. Assuming conditional independence of the approximate posterior as well, i.e. $q(\z | \x) = q(\z_0 | \x)q(\z_1 | \x) \cdots q(\z_k | \x)$, the Kullback-Leibler divergence simplifies as
\begin{align}
    &KL(q(\z|\x) || p(\z)) = \int_{\R^M} q(\z|\x) \log \dfrac{q(\z|\x)}{p(\z)} d\z \nonumber \\
    &= \int_{\R^{M_0} \times \cdots \times \R^{M_k}} q(\z_0 | \x)q(\z_1 | \x) \cdots q(\z_k | \x) \log \dfrac{q(\z_0 | \x)q(\z_1 | \x) \cdots q(\z_k | \x)}{p(\z_0)p(\z_1) \cdots p(\z_k)} d\z_0 \cdots d\z_k \nonumber \\
    &= \int_{R^{M_0}} q(\z_0|\x)\log \dfrac{q(\z_0)}{p(\z)} d\z_0 + \cdots + \int_{\R^{M_k}} q(\z_k | \x) \log \dfrac{q(\z_k | \x)}{p(\z_k)} d\z_k \nonumber \\
    &= \sum_i KL(q(\z_i | \x) || p(\z_i)),  \label{eq:dim-permutation-kl}
\end{align}

\section{Supplementary Tables and Figures} \label{ap:sup-tables-fig}

\begin{figure}[H]
    \centering
    \includegraphics[width=0.5\textwidth]{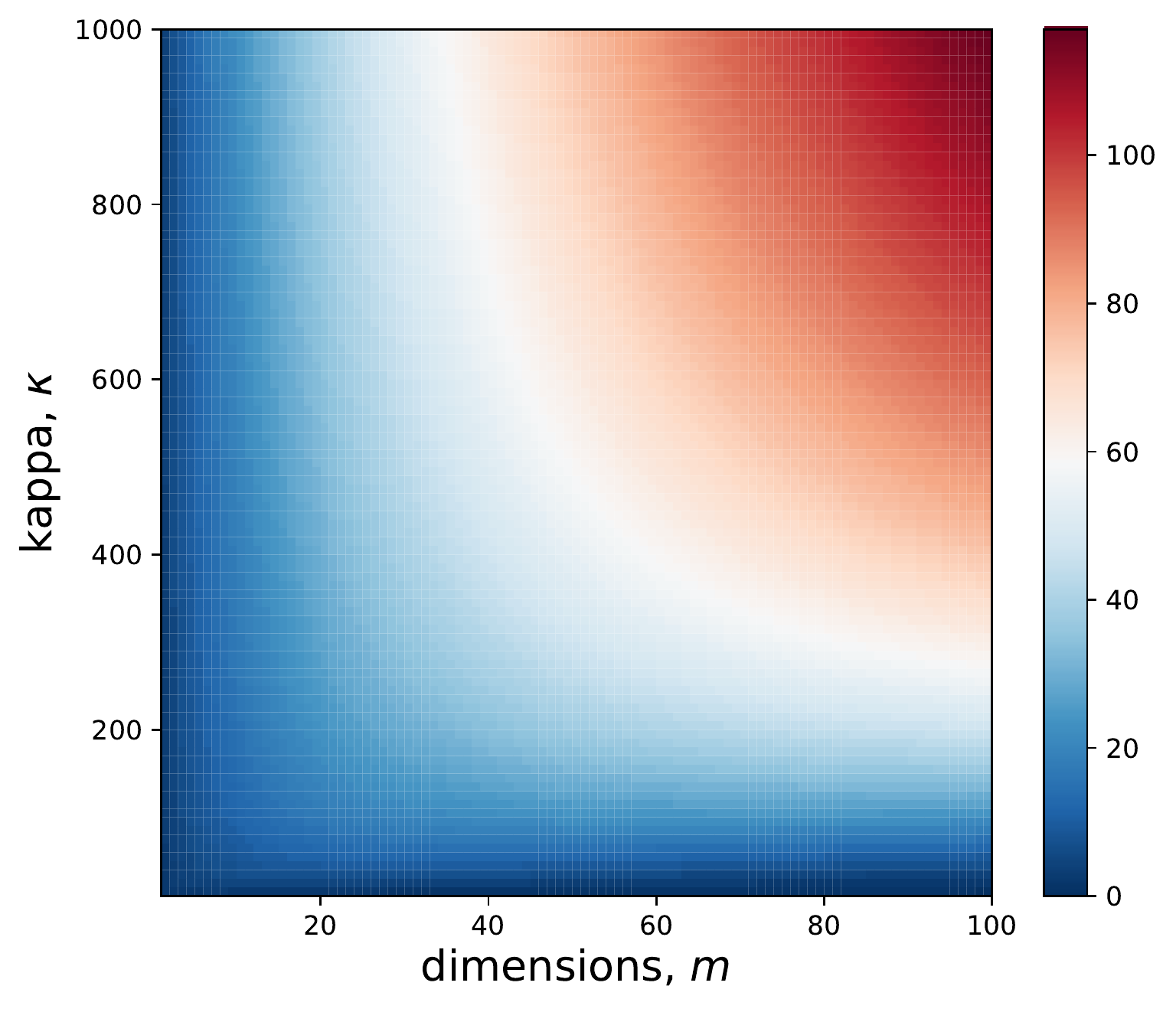}
    \caption{Value of the von Mises-Fisher Kullback-Leibler divergence varying the concentration parameter $\kappa$ on the y-axis, and the dimensionality $m$ on the x-axis. (Best viewed in color)}
    \label{fig:vmf-kl-divergence}
\end{figure}

Another way of understanding the importance of balance is by examining the KL divergence form of the vMF and its influence in the loss objective: In order to achieve high quality reconstruction performance, it is necessary for the concentration parameter $\kappa$ to concentrate, i.e. take on a high value. Given the Uniform prior setting in which $\kappa = 0$, this logically leads to an increase in the KL-divergence. The crucial observation here however, is that the strength of the KL-divergence is also strongly dependent on the dimensionality as can be observed in Fig. \ref{fig:vmf-kl-divergence}. Hence during learning over a product-space containing several lower dimensionality components and a single high dimensionality component, if the reconstruction error can be made sufficiently low using the lower dimensionality components, the optimal loss minimization strategy would be to set the concentration parameter of the largest component to 0, effectively ignoring it. A possible strategy to prevent this from happening could be to set separate $\beta$ parameters for each hyperspherical component, however we fear that this will quickly blow up the hyperparameter search-space.

\subsection{Fixed Ambient Space}

\begin{table}[H]
\centering
\caption{Summary of results of \Sm{40} ambient interpolations for unsupervised model on Static MNIST. RE and KL correspond respectively to the reconstruction and the KL part of the ELBO. 
}
\bigskip
  \bgroup
  \setlength\tabcolsep{.5em}
  \begin{tabular}{llccccc}
    \toprule
        m & $\kappa$ & [$\Sm{k}$] & LL & $\mathcal{L}|q|$ & $RE$ & $KL$ \\
        \midrule
        41 & 2 & $ \Sm{20} \times \Sm{19} $ & -93.18 & -98.72 & 69.78 & 28.94 \\ 
        41 & 2 & $ \Sm{38} \times \Sm{1} $ & -95.69 & -103.67 & 71.67 & 32.01 \\
        \midrule
        41 & 4 & $ \Sm{10} \times [\Sm{9}]_{\times 3} $ & -92.65 & -98.23 & 70.55 & 27.68 \\ 
        41 & 4 & $ \Sm{20 \times 10 \times 6 \times 1} $ & -92.59 & -98.27 & 71.33 & 26.94 \\ 
        41 & 4 & $ \Sm{34} \times [\Sm{1}]_{\times 3} $ & -108.42 & -116.86 & 99.62 & 17.23 \\ 
        \midrule
        41 & 6 & $ \Sm{15 \times 10 \times 4 \times 3 \times 2 \times 1} $ & -92.25 & -98.10 & 69.78 & 28.32 \\ 
        41 & 6 & $ \Sm{30} \times [\Sm{1}]_{\times 5} $ & -93.86 & -100.99 & 69.46 & 31.54 \\ 
        41 & 6 & $ [\Sm{6}]_{\times 5} \times \Sm{5} $ & -92.71 & -98.46 & 70.97 & 27.48 \\ 
        \midrule
        41 & 10 & $ \Sm{10 \times 5 \times 4 \times 3} \times [\Sm{2}]_{\times 3} \times [\Sm{1}]_{\times 3} $ & -92.93 & -99.07 & 70.67 & 28.41 \\ 
        41 & 10 & $ \Sm{22} \times [\Sm{1}]_{\times 9} $ & -93.45 & -100.29 & 68.75 & 31.54 \\ 
        41 & 10 & $ \Sm{4} \times [\Sm{3}]_{\times 9} $ & -93.36 & -99.40 & 71.93 & 27.47 \\ 
    \bottomrule
  \end{tabular}
  \egroup
  \label{tab:m-s40-ambient-static-unsupervised}
\end{table}

\begin{table}[H]
\centering
\caption{Summary of results of \Sm{40} ambient interpolations for unsupervised model on Caltech. 
}
\bigskip
  \bgroup
  \setlength\tabcolsep{.5em}
  \begin{tabular}{llccccc}
    \toprule
        m & $\kappa$ & [$\Sm{k}$] & LL & $\mathcal{L}|q|$ & $RE$ & $KL$ \\
        \midrule
        41 & 2 & $ \Sm{20} \times \Sm{19} $ & -142.43 & -155.24 & 123.35 & 31.89 \\ 
        41 & 2 & $ \Sm{38} \times \Sm{1} $ & -147.41 & -166.64 & 130.41 & 36.22 \\ 
        \midrule
        41 & 4 & $ \Sm{10} \times [\Sm{9}]_{\times 3} $ & -139.30 & -151.67 & 120.82 & 30.85 \\ 
        41 & 4 & $ \Sm{20 \times 10 \times 6 \times 1} $ & -140.64 & -153.05 & 123.23 & 29.82 \\ 
        41 & 4 & $ \Sm{34} \times [\Sm{1}]_{\times 3} $ & -168.25 & -186.47 & 170.44 & 16.03 \\ 
        \midrule
        41 & 6 & $\Sm{15 \times 10 \times 4 \times 3 \times 2 \times 1} $ & -142.84 & -156.59 & 126.59 & 30.00 \\ 
        41 & 6 & $ \Sm{30} \times [\Sm{1}]_{\times 5} $ & -169.15 & -177.23 & 161.68 & 15.55 \\ 
        41 & 6 & $ [\Sm{6}]_{\times 5} \times \Sm{5} $ & -139.99 & -152.68 & 121.91 & 30.77 \\ 
        \midrule
        41 & 10 & $ \Sm{10 \times 5 \times 4 \times 3} \times [\Sm{2}]_{\times 3} \times [\Sm{1}]_{\times 3} $ & -144.73 & -159.27 & 126.14 & 33.13 \\ 
        41 & 10 & $ \Sm{22} \times [\Sm{1}]_{\times 9} $ & -154.91 & -164.90 & 140.06 & 24.83 \\ 
        41 & 10 & $ \Sm{4} \times [\Sm{3}]_{\times 9} $ & -144.72 & -160.13 & 126.34 & 33.79 \\ 
    \bottomrule
  \end{tabular}
  \egroup
  \label{tab:m-s40-ambient-caltech-unsupervised}
\end{table}

\begin{table}[H]
\centering
\caption{Summary of results of \Sm{40} ambient interpolations for unsupervised model on Omniglot. 
}
\bigskip
  \bgroup
  \setlength\tabcolsep{.5em}
  \begin{tabular}{llccccc}
    \toprule
        m & $\kappa$ & [$\Sm{k}$] & LL & $\mathcal{L}|q|$ & $RE$ & $KL$ \\
        \midrule
        41 & 2 & $ \Sm{20} \times \Sm{19} $ & -114.32 & -120.72 & 92.10 & 28.62 \\ 
        41 & 2 & $ \Sm{38} \times \Sm{1} $ & -115.19 & -122.30 & 91.82 & 30.48 \\ 
        \midrule
        41 & 4 & $ \Sm{10} \times [\Sm{9}]_{\times 3} $ & -113.29 & -118.97 & 88.93 & 30.05 \\ 
        41 & 4 & $ \Sm{20 \times 10 \times 6 \times 1} $ & -112.79 & -119.17 & 87.94 & 31.23 \\ 
        41 & 4 & $ \Sm{34} \times [\Sm{1}]_{\times 3} $ & -136.39 & -142.03 & 132.75 & 9.28 \\ 
        \midrule
        41 & 6 & $ \Sm{15 \times 10 \times 4 \times 3 \times 2 \times 1} $ & -114.07 & -119.99 & 91.26 & 28.72 \\ 
        41 & 6 & $ \Sm{30} \times [\Sm{1}]_{\times 5} $ & -131.55 & -137.29 & 124.66 & 12.62 \\ 
        41 & 6 & $ [\Sm{6}]_{\times 5} \times \Sm{5} $ & -112.58 & -118.49 & 88.27 & 30.23 \\ 
        \midrule
        41 & 10 & $ \Sm{10 \times 5 \times 4 \times 3} \times [\Sm{2}]_{\times 3} \times [\Sm{1}]_{\times 3} $ & -113.53 & -119.83 & 90.00 & 29.83 \\ 
        41 & 10 & $ \Sm{22} \times [\Sm{1}]_{\times 9} $ & -114.95 & -121.42 & 92.42 & 29.00 \\ 
        41 & 10 & $ \Sm{4} \times [\Sm{3}]_{\times 9} $ & -112.64 & -118.67 & 88.98 & 29.68 \\ 
    \bottomrule
  \end{tabular}
  \egroup
  \label{tab:m-s40-ambient-omniglot-unsupervised}
\end{table}

\subsection{Fixed Degrees of Freedom}

\begin{table}[H]
\centering
\caption{ Overview of best results (mean over 3 runs) of \Sm{40} product-space interpolations compared to best single \Sm{m}-VAE ($m \le 40$) indicated (*). Here $a$ indicates the ambient space dimensionality, $\kappa$ the number of concentration parameters, i.e. breaks, and [\Sm{k}] the product-space composition.
}
\bigskip
  \bgroup
  \setlength\tabcolsep{.5em}
  \begin{tabular}{lcccccccc}
    \toprule
        \multirow[b]{2}{*}{\textbf{$a$}} & & & \multicolumn{5}{c}{Static MNIST} &
        \\
        & $\kappa$ & $[\mathcal{S}_k]$ & LL & $\mathcal{L}|q|$ & & LL* & $\mathcal{L}|q|$* \\
        \midrule
        44 & 4 & $ [\Sm{10}]_{\times 4} $ & -92.62 & -98.26 & & -96.32  & -104.11\\
        46 & 6 & $ [\Sm{7}]_{\times 5} \times \Sm{5} $ & -92.59 & -98.46 & & & \\
        46 & 6 & $ \Sm{15 \times 10 \times 5 \times 4 \times 3 \times 3} $ & -92.50 & -98.28 & & & \\
        50 & 10& $ \Sm{10 \times 7 \times 6 \times 5 \times 4 \times 3 \times 2} \times [\Sm{1}]_{\times 3} $ & -92.57 & -98.81 & & & \\
        \multicolumn{8}{c}{}\\
        & & & \multicolumn{5}{c}{Caltech} &  \\
        \midrule
        44 & 4 & $ [\Sm{10}]_{\times 4} $ & -137.95 & -150.86 & & -143.49  & -152.25 \\
        46 & 6 & $ [\Sm{7}]_{\times 5} \times \Sm{5} $ & -139.84 & -152.92 & & & \\
        \multicolumn{8}{c}{}\\
        & & & \multicolumn{5}{c}{Omniglot} & \\
        \midrule
        44 & 4 & $ [\Sm{10}]_{\times 4} $ & -112.28 & -118.21 & & -113.83 & -120.48 \\
        46 & 6 & $ [\Sm{7}]_{\times 5} \times \Sm{5} $ & -112.78 & -118.84 & & & \\
        50 & 10 & $ [\Sm{4}]_{\times 10} $ & -112.61 & -118.70 & & & \\
    \bottomrule
  \end{tabular}
  \egroup
  \label{tab:mtr-degree-s40-overview}
\end{table}

\subsection{Ignored and Disentangled Shells}

\begin{figure}[H]
\centering
     \subfigure[Ignored Sub-space]{\includegraphics[width=0.99\textwidth]{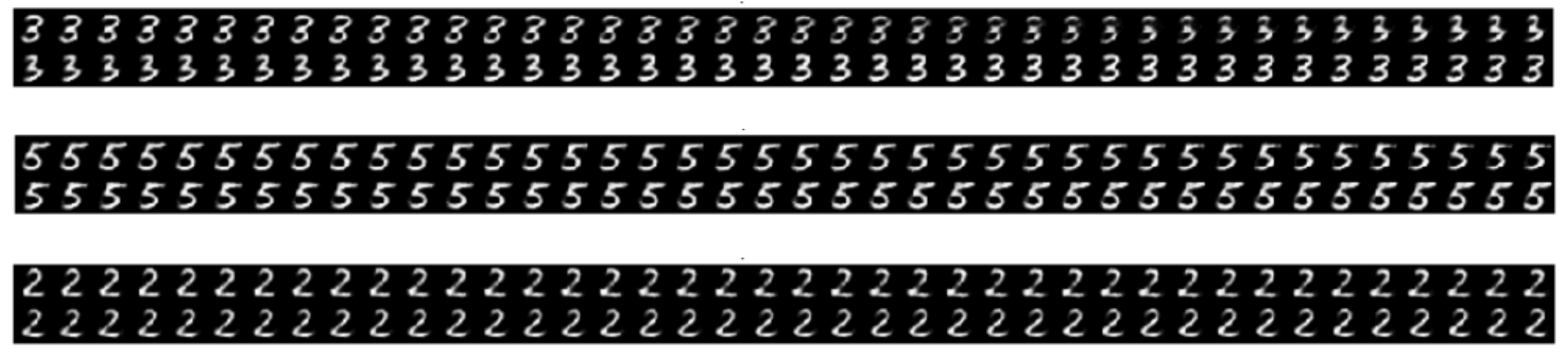} \label{fig:s9_1x9_ignored}
     } \hspace{2em}
     \subfigure[Thick to Thin]{\includegraphics[width=0.99\textwidth]{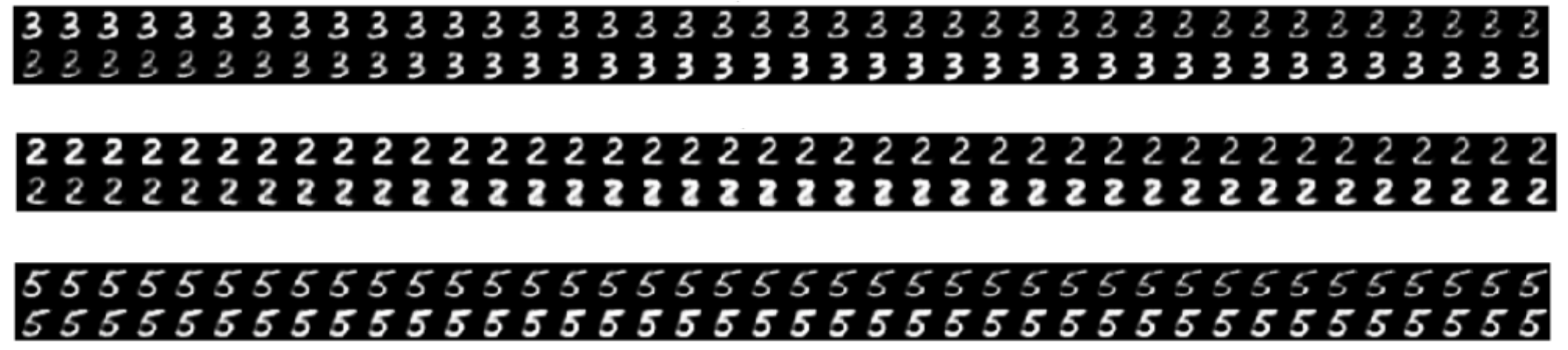} \label{fig:s9_1x9_thick_thin}
     }
\caption{\Sm{1} interpolations of broken up \Sm{9}. On top an example of an `ignored' sub-space, leading to little to no semantic change when decoded. Bottom a semantically meaningful sub-space that consistently changes the stroke thickness.}
\label{fig:s9_degree_mtr_interpolations}
\end{figure}

\end{document}